\documentclass{article}
\usepackage{spconf,amsmath,graphicx}

\usepackage{times}
\usepackage{epsfig}
\usepackage{booktabs}

\usepackage{amsmath,amssymb,amsfonts}
\usepackage{algorithmic}
\usepackage{graphicx}
\usepackage{textcomp}

\usepackage{framed,multirow}

\usepackage{amssymb}
\usepackage{latexsym}

\usepackage{url}
\usepackage{graphicx}
\usepackage{amsmath}
\usepackage{amssymb}
\usepackage{booktabs}
\usepackage{multicol}
\usepackage{makecell}
\usepackage{multirow}
\usepackage{subcaption}
\usepackage{blindtext}
\usepackage{cite}
\usepackage{tabularx}
\usepackage[pagebackref=true,breaklinks=true,letterpaper=true,colorlinks,bookmarks=false]{hyperref}

\title{DiffusionSTR: Diffusion Model for Scene Text Recognition}

\name{Masato Fujitake}
\address{FA Research, Fast Accounting Co., Ltd.
 \\
{\tt\small fujitake@fastaccounting.co.jp}
}

\begin{document}
\maketitle

\begin{abstract}
    This paper presents Diffusion Model for Scene Text Recognition  (DiffusionSTR), an end-to-end text recognition framework using diffusion models for recognizing text in the wild.
While existing studies have viewed the scene text recognition task as an image-to-text transformation, we rethought it as a text-text one under images in a diffusion model.
We show for the first time that the diffusion model can be applied to text recognition.
Furthermore, experimental results on publicly available datasets show that the proposed method achieves competitive accuracy compared to state-of-the-art methods.

\end{abstract}

\section{Introduction}
\label{sec:intro}
Text recognition in natural images is one of the active areas in computer vision and a fundamental and vital task in real-world applications such as document analysis and automated driving~\cite{fujitake2021tcbam, fujitake2023a3s}.
However, scene text recognition is challenging because it requires recognizing text in various fonts, colors, and shapes.
Many methods have been proposed to address this challenge.
Early research proposed methods that utilize information from images using Convolutional Neural Networks (CNNs) and recognizes text sequences using Recurrent Neural Networks (RNNs)~\cite{shi2016crnn}.
In order to deal with curved text images, pre-processing methods, such as Rectification Networks, were introduced before encoding visual features to improve accuracy~\cite{baek2021TRBA}.
Recently, methods with strong language models have been proposed for more robust recognition~\cite{fang2021ABINet, bautista2022parseq}.
While these methods contribute to high accuracy, they tend to be complex because they use multiple modules.

\begin{figure}[t]
    \centering
    \begin{subfigure}{.45\textwidth}
        \includegraphics[width=\linewidth]{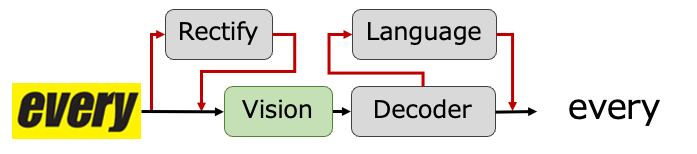}
        \caption{Typical approach of scene text recognition
        }
        \label{fig:recent_approach}
    \end{subfigure}\hfill \\
    \vspace*{1.00\baselineskip}
    \begin{subfigure}{.45\textwidth}
        \includegraphics[width=\linewidth]{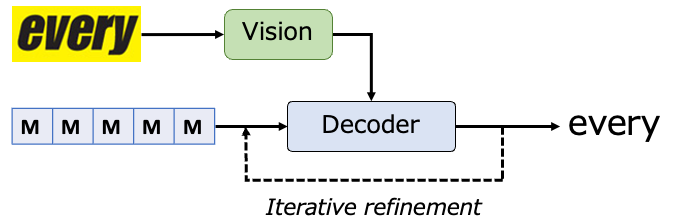}
        \caption{Our approach using diffusion model
        }
        \label{fig:our_approach}
    \end{subfigure}\hfill
    \caption{
    \textbf{The overview of the typical approach and our proposed one.}
Typical approaches consist of a vision module to obtain information from an image and a decoder to convert it to a text sequence.
Some methods employ rectified modules and language models to boost accuracy, which leads to complex architecture.
In contrast, our approach consists of two main components.
The decoder converts a sequence to a recognized result iteratively under the visual condition.
    }
    \label{fig:abstract_approach}
    \vspace*{-1.00\baselineskip}
\end{figure}

We propose a new approach by reviewing the input-output relationship of the scene text recognition task.
While existing approaches generate text sequences directly from images, as shown in Figure~\ref{fig:recent_approach}, our method iteratively transforms a text sequence into a correctly recognized one under the condition of image information, as shown in Figure~\ref{fig:our_approach}.
In other words, we regard the text recognition task as a text-to-text transformation task rather than an image-to-text task.
By matching the input-output relationship of the same domain, rather than directly converting the different domains of image and text, prediction is possible even in simple structures, unlike existing methods.

We introduce a diffusion model into the scene text recognition task to realize our approach.
In recent years, generative models based on diffusion models have succeeded in image generation~\cite{ho2020ddpm, Hoogeboom2021argmax}.
The probabilistic diffusion models pipeline consists of a chain of Markov latent variables, with data flowing in two directions: the diffusion process and the denoising process. 
The denoising process generates data from Gaussian noise through inference.
In contrast, the diffusion process is the training process that learns to transform data samples into Gaussian noise.
We leverage a diffusion model in our proposed approach.
The main structural feature of the diffusion model in the image generation task is that the input-output relationship corresponds to the same resolution---a fixed dimension---for images.
Unlike image generation, however, the dimension of scene text recognition varies from image to image due to the length of a text sequence.
This makes learning difficult because there is a different challenge of where a text ends within a fixed-length sequence in addition to the categorical classification of characters.
To solve this problem, we propose a character-aware head, which predicts whether a character exists at the position in the sequence.
By doing so, we achieved accuracy comparable to leading competitive methods despite its simple structure.

Our contribution is listed as follows.
\begin{itemize}
    \item We propose DiffusionSTR, the first framework for scene text recognition with diffusion models.
    \item Experimental results show the proposed method's effectiveness, achieving competitive accuracy with state-of-the-art methods on public datasets.
\end{itemize}

\section{Related Works} \noindent
\textbf{Scene Text Recognition.}
Scene Text Recognition can be roughly divided into two categories: language-free and language-based approaches.
The language-free approaches predict the sequence of a character directly from input images without any language constraint.
The main methods are CTC-based and segmentation-based methods.
The CTC-based methods~\cite{hu2020gtc, shi2016crnn} combine CNN to extract visual features and sequence models, such as RNN, to predict a sequence of characters with end-to-end training using CTC loss~\cite{graves2006ctcloss}.
The segmentation-based methods segment characters at pixel level and recognize them by grouping~\cite{wan2020textscanner}.
However, these approaches do not use linguistic information, only image information, making them vulnerable to noise, such as occlusion and distortion.

The language-based approaches have been studied recently to alleviate the above problems~\cite{fang2021ABINet}.
Early research proposed methods utilizing N-grams~\cite{jaderberg2014MJSynth}, but recent methods have been proposed using powerful language models such as those represented by RNNs~\cite{shi2018aster} and Transformer~\cite{fang2021ABINet}.
For instance, ABINet inputs the recognition results from vision into a language model and obtains text results with added language information. 
Then, it predicts the refined results fusing the two recognition results.
This process is iterative for improvement~\cite{fang2021ABINet}.
Moreover, PARSeq uses Permutation Language Modeling to sophisticate the iterative improvement process~\cite{bautista2022parseq}.
These methods have contributed to higher accuracy by introducing powerful language models, but they are complex mechanisms.

Our proposed method differs from existing research in two ways.
First, we do not use a language model for simplicity.
Second, while existing research performs direct inference from images to text results, our proposed method differs because we prepare token sequences in advance and transform them to correct sequences through the vision condition.

\noindent
\textbf{Diffusion Models.}
Diffusion models are latent variable generative frameworks in~\cite{sohl2015diffusionmodel} and are improved in~\cite{ho2020ddpm}.
Due to their remarkable power, they have recently attracted attention in image generation~\cite{sohl2015diffusionmodel, ho2020ddpm,ramesh2022hierarchical}, mainly in continuous data.
Adaptations have also been proposed for object detection~\cite{chen2022diffusiondet} and audio~\cite{kong2021diffwave}.
Although initially not directly applicable to discrete data such as text, the extension to the discrete data has recently been proposed~\cite{Hoogeboom2021argmax}.
However, it has yet to be shown to what extent the diffusion model is effective in text recognition tasks.
This is the first work of the diffusion models for the text recognition task.
\label{sec:relatedwork}

\section{Method} \label{sec:method}

\begin{figure*}[htp]
	\centering
	\includegraphics[width=1.40\columnwidth, keepaspectratio]{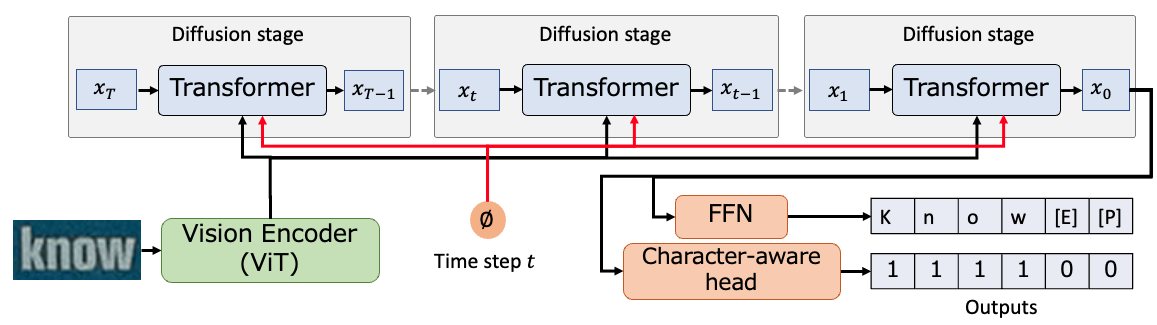}
	\caption{
\textbf{The pipeline of the proposed scene text recognition using a diffusion model.}
It consists of the vision encoder~\cite{dosovitskiy2020vit}, Transformer~\cite{vaswani2017transformer} with an additional time-based positional encoding, FFN and character-aware head.
\texttt{[E]} and \texttt{[P]} are abbreviations for special tokens \texttt{[EOS]} and \texttt{[PAD]}, respectively.
The proposed method performs text recognition by repeatedly refining the input sequence $x_{t}$ based on image information.
FFN performs character classification, while the character-aware head predicts where a character exists in a fixed-length sequence.
	}
	\label{fig:overall_architecture}
    \vspace*{-1.00\baselineskip}
\end{figure*}

Figure~\ref{fig:overall_architecture} shows the pipeline of the proposed method.
Our proposed method learns to transform text-to-text for scene text recognition through a diffusion model process based on the transformer architecture, including vision and text.
The model comprises a vision encoder, a transformer, linear layers---FFN, and a character-aware head---that transform the results.
The overall flow begins with generating visual features from images using the vision encoder.
Next, a noise-filled token sequence $x_{T}$ is used as input to generate a refined one $x_{T-1}$ through the Transformer~\cite{vaswani2017transformer} under visual feature conditions.
The new token sequence is refined $T$ times, and finally, the output $x_{0}$ is converted to the recognized text through FFN, and the character's position is predicted through the character-aware head.
We describe each detail below.

\subsection{Diffusion Model}
The original diffusion model~\cite{sohl2015diffusionmodel} constructs forward and reverse processes.
The forward process gradually deteriorates a data point $x_{0}$, sampled from a real-world data distribution $x_{0} \sim q(x)$.
It adds a small amount of noise at each time step where $t \in 1, \cdots T$ and makes the data point into a Gaussian noise $x_{T} \sim \mathcal{N}(0, \mathbf{I})$.
On the other hand, the reverse denoising process tries to gradually reconstruct the original data $x_{0}$ through sampling from $x_{T}$.
It is described as a learnable distribution $p(x_{t-1} \mid x_{t})$. 

In applying a diffusion model to the text recognition task, we basically follow the diffusion models~\cite{sohl2015diffusionmodel, Hoogeboom2021argmax} but with some modifications.
First, in the image generation task, the data $x_{t}$ corresponds to a single image. 
However, our scene text recognition corresponds to a single text sequence comprising character tokens.
Moreover, we use the multinomial diffusion model~\cite{Hoogeboom2021argmax} for categorical data because, unlike the image generation task whose variable is continuance, scene text recognition's variable is discrete.
In addition to characters, we also use special tokens to correspond to text recognition: the \texttt{[EOS]} token indicates the end of the recognized text. 
The \texttt{[PAD]} token indicates the padding of the fixed-length sequence after \texttt{[EOS]}.
The \texttt{[MASK]} token is a special token that signifies noise state.

Next, although the original diffusion model is optimized to maximize the marginal likelihood of the data in~\cite{sohl2015diffusionmodel}, we use the devised objective $L_{simple}$~\cite{ho2020ddpm} with a mean-squared error loss for stable training.

\subsection{Vision Encoder}
The vision encoder $VisEnc$ extracts information from a given image.
For this purpose, we used ViT~\cite{dosovitskiy2020vit}, an extension of Transformer~\cite{vaswani2017transformer} to the image field.
That is, image $I \in \mathbb{R}^{H\times W \times C}$ is tokenized per $p_{w} \times p_{h}$ patch and encoded with 12 ViT-layer, where $H, W, C$ are the height, width, and channel of the image.
We used a learned positional encoding.
Therefore, the visual features $z$ are defined, 

\begin{equation}\label{Eq:vision_encoder}
   z = \mathrm{VisEnc}(I) \in {\mathbb{R}^{{{ \frac{HW}{p_{w}p_{h}} \times d}}}},
\end{equation}
where $d$ is the embedding dimension of ViT.

\subsection{Transformer}
We employ the Transformer decoder~\cite{vaswani2017transformer} $Dec$ with an additional time-based positional encoding $\phi$ to convert token sequences in diffusion models.
The architecture is similar to the decoder in the unconditional text translation with a diffusion model~\cite{Hoogeboom2021argmax}.
However, it differs in three respects.
The first is the text transformation under the condition of the vision features $z$.
We use the Transformer's cross-attention mechanism within the decoder to condition the text sequence.
Second, while the Transformer typically infers one token at a time to handle text, our text recognition task simultaneously outputs the probabilities of all tokens.
Third, the Transformer output results are used in two ways: one is converted into a string for text recognition using the Feedforward Network (FFN), following previous research~\cite{bautista2022parseq}; the other is input into a character-aware head, which has an identical structure to the FFN, to predict whether a character region contains a real character, such as the alphabet.
We use the cross-entropy loss for FFN.

The time positional encoding utilizes sinusoidal positional embedding and linear layers. The output is added to the typical sequence positional encoding in the Transformer and put into each decoder layer.

\subsection{Character-aware Head}
We propose a character-aware head for text recognition in the diffusion model.
Since text recognition in the diffusion model is performed within a fixed-length sequence, there are two issues: the categorical classification of characters and the extent to which a character is a character sequence.
Therefore, we propose classifying at a large level what is a character domain in a fixed-length sequence and facilitating categorical classification of characters.
As shown in the output of Figure~\ref{fig:overall_architecture}, the character-aware head performs a binary classification of whether the position corresponds to a character (1) or not (0).
Binary Cross-entropy was used as the loss function.

\subsection{Training Protocol}
During training, we randomly sample a time step $t$.
Then, we calculate the posteriors using the noise schedule $\alpha_{t}$, $\bar{\alpha}_{t}$, as defined in the original diffusion model, and a noisy sequence $x_{t}$ as following~\cite{Hoogeboom2021argmax}.
We perform $x_{t-1} = Dec(x_{t}, z, t)$ to compute the loss.

\subsection{Inference Protocol}
The Inference process starts with the sequence $x_{T}$, filled with a mask token.
The process is conditioned on $z$ from the vision encoder.
It then iterates $T$ times to obtain $x_{0}$.
We convert the final token sequence to text with FFN.

\section{Experiments} \label{sec:experiments}

\subsection{Dataset and Evaluation}
For a fair comparison, we conducted experiments following the setup of~\cite{yu2020srn}.
We train the models on two synthetic datasets MJSynth (MJ)~\cite{jaderberg2014MJSynth, jaderberg2016MJSynth} and SynthText (ST)~\cite{gupta2016synthtext}.
We evaluated models on six standard benchmarks: ICDAR 2013 (IC13)~\cite{karatzas2013icdar}, ICDAR 2015 (IC15)~\cite{karatzas2015icdar}, IIIT 5KWords (IIIT)~\cite{mishra2012iiit}, Street View Text (SVT)~\cite{wang2011svt}, Street View Text-Perspective (SVTP)~\cite{phan2013svtp} and CUTE80 (CUTE)~\cite{risnumawan2014cute80}.
For evaluation, we use word-level accuracy on the six benchmark datasets.
A prediction is considered correct if characters at all positions match.
We report the mean score of the four experiments by following the previous research~\cite{bautista2022parseq}.

\subsection{Implementation Details}
All experiments are performed on four Nvidia A100 GPUs in mixed precision using PyTorch.
We used the ADAMW optimizer~\cite{loshchilov2018adamw} with a learning rate, which warms up to $10^{-4}$ linearly and drops to 0 following cosine decay.
We set the hyperparameters of the optimizer $\epsilon$ and $\beta$ as $10^{-8}$ and (0.9, 0.999), respectively.
The number of training epochs is 20, and the warm-up epoch is 5.
The batch size is 384.
We set the weight decay to 0.01
For the Label preprocessing, we follow the previous work~\cite{shi2018aster}. Concretely, we set a maximum label length of 25 and a charset size of 94, which includes mixed-case alphanumeric characters and punctuation marks without special tokens.
We strictly follow the way of image preprocessing~\cite{bautista2022parseq} including augmentation.
For a fair comparison, we used a vision encoder~\cite{dosovitskiy2020vit} with the same settings as PARSeq~\cite{bautista2022parseq}.
We have six transformer layers with 16 attention heads and hidden dimension $d$ of 384 for the transformer decoder.
The balanced weights of the cross entropy loss function for character-aware head and character recognition are equal.
We set the time step $T$ to 1000.

\begin{table}[b]

\centering
   \caption{
\textbf{Word accuracy on the six benchmark datasets.}
}
\label{tab:method_overall_result}
\resizebox{1.0\columnwidth}{!}{%
\begin{tabular}{ccccccccc}
\hline
\multirow{3}{*}{Method}  & \multicolumn{8}{c}{Test datasets and \# of samples}                                                                                                                                                                   \\ \cline{2-9} 
                         & IIIT5k                   & SVT                      & \multicolumn{2}{c}{IC13}                            & \multicolumn{2}{c}{IC15}                            & SVTP                     & CUTE                     \\ 
                         & 3,000                    & 647                      & 857                      & 1,015                    & 1,811                    & 2,077                    & 645                      & 288                      \\ \hline
CRNN~\cite{shi2016crnn} & 81.8 & 80.1 & 89.4 & 88.4 & 65.3 & 60.4 & 65.9 & 61.5 \\ 
ViTSTR~\cite{atienza2021vitstr} & 88.4 & 87.7 & 93.2 & 92.4 & 78.5 & 72.6 & 81.8 & 81.3 \\
TRBA~\cite{baek2021TRBA} & 92.1 & 88.9 & $-$ & 93.1 & $-$ & 74.7 & 79.5 & 78.2 \\
ABINet~\cite{fang2021ABINet} & 96.2 & 93.5 & \textbf{97.4} & $-$ & 86.0 & $-$ & \textbf{89.3} & 89.2 \\
PARSeq~\cite{bautista2022parseq} & 97.0 & \textbf{93.6} & 97.0 & 96.2 & \textbf{86.5} & \textbf{82.9} & 88.9 & 92.2 \\
\midrule

DiffusionSTR (Ours) & \textbf{97.3} & \textbf{93.6} & 97.1 & \textbf{96.4} & 86.0 & 82.2 & 89.2 & \textbf{92.5} \\

\hline

\end{tabular}
}
    \vspace*{-1.00\baselineskip}
\end{table}

\subsection{Main Results}

In this section, we compare the performances of the proposed method with state-of-the-art methods on public datasets.
Table~\ref{tab:method_overall_result} shows the results of state-of-the-art methods and our experiments.
For a fair comparison, we list the methods with only MJ and ST datasets for training.
Our method has reached competitive accuracy against the latest strong methods and outperforms them on several datasets.
ABINet~\cite{fang2021ABINet} and PARSeq~\cite{bautista2022parseq} use powerful language models, while TRBA~\cite{baek2021TRBA} uses an image rectification module as preprocessing.
Against such models, our model achieves comparable accuracy despite its simple structure without using either, demonstrating the effectiveness of the proposed method.

\begin{figure}[htp]
	\centering
	\includegraphics[width=0.90\columnwidth, keepaspectratio]{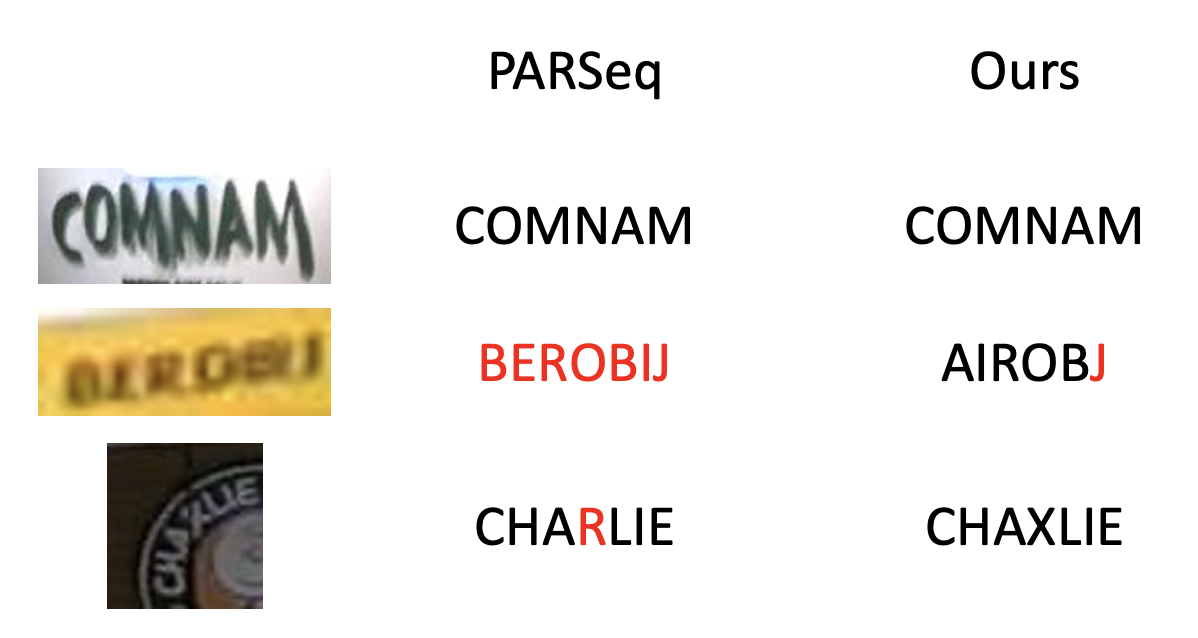}
	\caption{
\textbf{Visualized comparison against the state-of-the-art method~\cite{bautista2022parseq}.}
The leftmost column shows the input image, and each column displays each method's output.
Black text indicates a match with the ground truth; red indicates a different case.
Our method recognizes more robustly.
	}
	\label{fig:visualized_comparison}
    \vspace*{-0.30\baselineskip}
\end{figure}

\subsection{Detailed Analysis}
To confirm the effectiveness of the proposed method in detail, we conducted a detailed analysis.
In this section, we report the average accuracy of all test sets.

\noindent
\textbf{Qualitative analysis.}
Figure~\ref{fig:visualized_comparison} shows the text recognition results of the proposed method and the latest one using the language model~\cite{bautista2022parseq}.
Our method outputs reasonable results, even for blurred or curved images.
It produces more appropriate results for noisy images than the existing one.

\begin{table}[bt]
\centering
\caption{
	\textbf{Impact of character-aware head.}
}
\label{tab:ablation_characterawarehead}
\resizebox{0.8\columnwidth}{!}{
\begin{tabular}{c|c|c}
\toprule
Model                                & Character-aware head & Average accuracy    \\ \midrule
Complete model & \checkmark &  91.8   \\
Ablation model & $-$ &  63.4   \\
\bottomrule
\end{tabular}
}
\vspace*{-1.00\baselineskip}
\end{table}

\noindent
\textbf{Impact of character-aware head.}
In addition to the categorical classification of characters, DiffusionSTR proposes to predict the position of character presence.
Table~\ref{tab:ablation_characterawarehead} shows the effectiveness of the proposed character-aware head.
We see that the accuracy without predicting the presence is significantly degraded.
The diffusion model needs to infer the location of character presence for text recognition.

\begin{table}[tb]
    \centering
    \caption{
    \textbf{Impact of the time step $T$. }
    }
    \vspace*{-0.75\baselineskip}
    \label{tab:ablation_result_of_n_steps}
    \resizebox{0.80\columnwidth}{!}{%
        \begin{tabular}{l|cccccc} \toprule
        Total  step ($T$) &  100    & 500 & 1000  & 2000   & 4000    \\ \midrule 
       Average accuracy   &  17.6   & 86.1 & 91.8  & 91.9   & 91.3  \\ 
    \bottomrule
    \end{tabular}
    }
    \vspace*{-1.00\baselineskip}
\end{table}

\noindent
\textbf{Impact of time step.}
The diffusion step is a vital factor in the diffusion model.
Table~\ref{tab:ablation_result_of_n_steps} shows how the total number of steps affects accuracy.
We confirm that the accuracy increases as the number of diffusion steps increases; after 1000, only a little difference can be observed.

\section{Conclusion}\label{sec:conclusion}
This work proposed a novel scene text recognition framework using diffusion models, which refines a noisy text sequence to the recognized one iteratively.
Unlike existing methods that directly transform visual information into a text sequence, the proposed method leverages visual information to refine noisy text sequences conditionally.
Experiment results showed that our method achieved competitive results compared to state-of-the-art methods with simple architecture.

\clearpage

{\small
\bibliographystyle{IEEEbib}
\bibliography{article}

\begin{thebibliography}{10}

\bibitem{fujitake2021tcbam}
Masato Fujitake and Hongpeng Ge,
\newblock ``Temporally-aware convolutional block attention module for video
  text detection,''
\newblock in {\em IEEE SMC}, 2021, pp. 220--225.

\bibitem{fujitake2023a3s}
Masato Fujitake,
\newblock ``A3s: Adversarial learning of semantic representations for
  scene-text spotting,''
\newblock in {\em ICASSP}, 2023, pp. 1--5.

\bibitem{shi2016crnn}
Baoguang Shi, Xiang Bai, and Cong Yao,
\newblock ``An end-to-end trainable neural network for image-based sequence
  recognition and its application to scene text recognition,''
\newblock {\em TPAMI}, vol. 39, no. 11, pp. 2298--2304, 2016.

\bibitem{baek2021TRBA}
Jeonghun Baek, Yusuke Matsui, and Kiyoharu Aizawa,
\newblock ``What if we only use real datasets for scene text recognition?
  toward scene text recognition with fewer labels,''
\newblock in {\em CVPR}, 2021, pp. 3113--3122.

\bibitem{fang2021ABINet}
Shancheng Fang, Hongtao Xie, Yuxin Wang, Zhendong Mao, and Yongdong Zhang,
\newblock ``Read like humans: Autonomous, bidirectional and iterative language
  modeling for scene text recognition,''
\newblock in {\em CVPR}, 2021, pp. 7098--7107.

\bibitem{bautista2022parseq}
Darwin Bautista and Rowel Atienza,
\newblock ``Scene text recognition with permuted autoregressive sequence
  models,''
\newblock in {\em ECCV}, 2022, pp. 178--196.

\bibitem{ho2020ddpm}
Jonathan Ho, Ajay Jain, and Pieter Abbeel,
\newblock ``Denoising diffusion probabilistic models,''
\newblock in {\em NeurIPS}, 2020, vol.~33, pp. 6840--6851.

\bibitem{Hoogeboom2021argmax}
Emiel Hoogeboom, Didrik Nielsen, Priyank Jaini, Patrick Forr\'{e}, and Max
  Welling,
\newblock ``Argmax flows and multinomial diffusion: Learning categorical
  distributions,''
\newblock in {\em NeurIPS}, 2021, vol.~34, pp. 12454--12465.

\bibitem{hu2020gtc}
Wenyang Hu, Xiaocong Cai, Jun Hou, Shuai Yi, and Zhiping Lin,
\newblock ``Gtc: Guided training of ctc towards efficient and accurate scene
  text recognition,''
\newblock in {\em AAAI}, 2020, vol.~34, pp. 11005--11012.

\bibitem{graves2006ctcloss}
Alex Graves, Santiago Fern{\'a}ndez, Faustino Gomez, and J{\"u}rgen
  Schmidhuber,
\newblock ``Connectionist temporal classification: labelling unsegmented
  sequence data with recurrent neural networks,''
\newblock in {\em ICML}, 2006, pp. 369--376.

\bibitem{wan2020textscanner}
Zhaoyi Wan, Minghang He, Haoran Chen, Xiang Bai, and Cong Yao,
\newblock ``Textscanner: Reading characters in order for robust scene text
  recognition,''
\newblock in {\em AAAI}, 2020, vol.~34, pp. 12120--12127.

\bibitem{jaderberg2014MJSynth}
Max Jaderberg, Karen Simonyan, Andrea Vedaldi, and Andrew Zisserman,
\newblock ``Synthetic data and artificial neural networks for natural scene
  text recognition,''
\newblock in {\em NIPS Workshop}, 2014.

\bibitem{shi2018aster}
Baoguang Shi, Mingkun Yang, Xinggang Wang, Pengyuan Lyu, Cong Yao, and Xiang
  Bai,
\newblock ``Aster: An attentional scene text recognizer with flexible
  rectification,''
\newblock {\em TPAMI}, vol. 41, no. 9, pp. 2035--2048, 2018.

\bibitem{sohl2015diffusionmodel}
Jascha Sohl-Dickstein, Eric Weiss, Niru Maheswaranathan, and Surya Ganguli,
\newblock ``Deep unsupervised learning using nonequilibrium thermodynamics,''
\newblock in {\em ICML}. PMLR, 2015, pp. 2256--2265.

\bibitem{ramesh2022hierarchical}
Aditya Ramesh, Prafulla Dhariwal, Alex Nichol, Casey Chu, and Mark Chen,
\newblock ``Hierarchical text-conditional image generation with clip latents,''
\newblock {\em arXiv preprint arXiv:2204.06125}, 2022.

\bibitem{chen2022diffusiondet}
Shoufa Chen, Peize Sun, Yibing Song, and Ping Luo,
\newblock ``Diffusiondet: Diffusion model for object detection,''
\newblock {\em arXiv preprint arXiv:2211.09788}, 2022.

\bibitem{kong2021diffwave}
Zhifeng Kong, Wei Ping, Jiaji Huang, Kexin Zhao, and Bryan Catanzaro,
\newblock ``Diffwave: A versatile diffusion model for audio synthesis,''
\newblock in {\em ICLR}, 2021.

\bibitem{dosovitskiy2020vit}
Alexey Dosovitskiy, Lucas Beyer, Alexander Kolesnikov, Dirk Weissenborn,
  Xiaohua Zhai, Thomas Unterthiner, Mostafa Dehghani, Matthias Minderer, Georg
  Heigold, Sylvain Gelly, et~al.,
\newblock ``An image is worth 16x16 words: Transformers for image recognition
  at scale,''
\newblock in {\em ICLR}, 2020.

\bibitem{vaswani2017transformer}
Ashish Vaswani, Noam Shazeer, Niki Parmar, Jakob Uszkoreit, Llion Jones,
  Aidan~N Gomez, \L~ukasz Kaiser, and Illia Polosukhin,
\newblock ``Attention is all you need,''
\newblock in {\em NIPS}, 2017, pp. 5998--6008.

\bibitem{yu2020srn}
Deli Yu, Xuan Li, Chengquan Zhang, Tao Liu, Junyu Han, Jingtuo Liu, and Errui
  Ding,
\newblock ``Towards accurate scene text recognition with semantic reasoning
  networks,''
\newblock in {\em CVPR}, 2020, pp. 12113--12122.

\bibitem{jaderberg2016MJSynth}
Max Jaderberg, Karen Simonyan, Andrea Vedaldi, and Andrew Zisserman,
\newblock ``Reading text in the wild with convolutional neural networks,''
\newblock {\em IJCV}, vol. 116, pp. 1--20, 2016.

\bibitem{gupta2016synthtext}
Ankush Gupta, Andrea Vedaldi, and Andrew Zisserman,
\newblock ``Synthetic data for text localisation in natural images,''
\newblock in {\em CVPR}, 2016, pp. 2315--2324.

\bibitem{karatzas2013icdar}
Dimosthenis Karatzas, Faisal Shafait, Seiichi Uchida, Masakazu Iwamura,
  Lluis~Gomez i~Bigorda, Sergi~Robles Mestre, Joan Mas, David~Fernandez Mota,
  Jon~Almazan Almazan, and Lluis~Pere De~Las~Heras,
\newblock ``Icdar 2013 robust reading competition,''
\newblock in {\em ICDAR}, 2013, pp. 1484--1493.

\bibitem{karatzas2015icdar}
Dimosthenis Karatzas, Lluis Gomez-Bigorda, Anguelos Nicolaou, Suman Ghosh,
  Andrew Bagdanov, Masakazu Iwamura, Jiri Matas, Lukas Neumann,
  Vijay~Ramaseshan Chandrasekhar, Shijian Lu, et~al.,
\newblock ``Icdar 2015 competition on robust reading,''
\newblock in {\em ICDAR}, 2015, pp. 1156--1160.

\bibitem{mishra2012iiit}
Anand Mishra, Karteek Alahari, and CV~Jawahar,
\newblock ``Scene text recognition using higher order language priors,''
\newblock in {\em BMVC}, 2012.

\bibitem{wang2011svt}
Kai Wang, Boris Babenko, and Serge Belongie,
\newblock ``End-to-end scene text recognition,''
\newblock in {\em ICCV}, 2011, pp. 1457--1464.

\bibitem{phan2013svtp}
Trung~Quy Phan, Palaiahnakote Shivakumara, Shangxuan Tian, and Chew~Lim Tan,
\newblock ``Recognizing text with perspective distortion in natural scenes,''
\newblock in {\em ICCV}, 2013, pp. 569--576.

\bibitem{risnumawan2014cute80}
Anhar Risnumawan, Palaiahankote Shivakumara, Chee~Seng Chan, and Chew~Lim Tan,
\newblock ``A robust arbitrary text detection system for natural scene
  images,''
\newblock {\em Expert Systems with Applications}, vol. 41, no. 18, pp.
  8027--8048, 2014.

\bibitem{loshchilov2018adamw}
Ilya Loshchilov and Frank Hutter,
\newblock ``Decoupled weight decay regularization,''
\newblock in {\em ICLR}, 2018, pp. 1--10.

\bibitem{atienza2021vitstr}
Rowel Atienza,
\newblock ``Vision transformer for fast and efficient scene text recognition,''
\newblock in {\em ICDAR}, 2021, pp. 319--334.

\end{thebibliography}
}

\end{document}